\documentclass[11pt,a4paper]{article}
\usepackage[hyperref]{emnlp2020}
\usepackage{times}
\usepackage{latexsym}
\usepackage{hyperref}
\usepackage{url}
\usepackage{times}
\usepackage{latexsym}
\usepackage{times}
\usepackage{epsfig}
\usepackage{graphicx}
\usepackage{amsmath}
\usepackage{amssymb}
\usepackage{color}
\usepackage{amsmath}
\usepackage{amssymb}
\usepackage{fancyhdr}
\usepackage{listings}
\usepackage{multirow}
\usepackage{graphicx}
\usepackage{verbatim}
\usepackage{bbm}
\usepackage{color}
\usepackage{mathtools}
\usepackage{pifont}
\usepackage{makecell}
\usepackage{multirow}
\usepackage{booktabs}
\usepackage{float}
\usepackage{amsmath,amsfonts,amsthm,bm}
\usepackage{enumitem}
\usepackage{xurl}
\usepackage{pifont}
\usepackage{xcolor}
\usepackage{sidecap}
\usepackage{booktabs,siunitx} 
\usepackage{tikz}
\usepackage{array}
\usepackage{algorithmic}
\usepackage{algorithm}
\usepackage{subfig}

\usepackage{microtype}

\aclfinalcopy 

\title{Efficient Meta Lifelong-Learning with Limited Memory}

\author{Zirui Wang\thanks{\hspace{1mm} Equal contribution, name order decided by coin flip.}, Sanket Vaibhav Mehta\footnotemark[1], Barnab\'{a}s P\'{o}czos, Jaime Carbonell \\
 Carnegie Mellon University, Pittsburgh, USA \\
 {\texttt{\{ziruiw, svmehta, bapoczos, jgc\}@cs.cmu.edu}} }

\date{}

\begin{document}
\maketitle
\begin{abstract}
Current natural language processing models work well on a single task, yet they often fail to continuously learn new tasks without forgetting previous ones as they are re-trained  throughout their lifetime, a challenge known as lifelong learning.
State-of-the-art lifelong language learning methods store past examples in episodic memory and replay them at both training and inference time.
However, as we show later in our experiments, there are three significant impediments: 
(1) needing unrealistically large memory module to achieve good performance, 
(2) suffering from negative transfer, 
(3) requiring multiple local adaptation steps for each test example that significantly slows down the inference speed.
In this paper, we identify three common principles of lifelong learning methods and propose an efficient meta-lifelong framework that combines them in a synergistic fashion.
To achieve sample efficiency, our method trains the model in a manner that it learns a better initialization for local adaptation.
Extensive experiments on text classification and question answering benchmarks demonstrate the effectiveness of our framework by achieving state-of-the-art performance using merely 1\% memory size and narrowing the gap with multi-task learning.
We further show that our method alleviates both catastrophic forgetting and negative transfer at the same time.
\end{abstract}

\section{Introduction}

Humans learn throughout their lifetime, 
 quickly adapting to new environments and acquiring new skills by leveraging past experiences, while retaining old skills and continuously accumulating knowledge.
However, state-of-the-art machine learning models rely on the data distribution being stationary and struggle in learning diverse tasks in such a \textit{lifelong learning} setting \citep{parisi2019continual} (see section \ref{sec:background} for a formal definition).
In particular, they fail to either effectively reuse previously acquired knowledge to help learn new tasks, or they forget prior skills when learning new ones - these two phenomena are known as \textit{negative transfer} \citep{wang2019characterizing} and \textit{catastrophic forgetting} \citep{mccloskey1989catastrophic}, respectively. 
These downsides limit applications of existing models to real-world environments that dynamically evolve.

Due to its potential practical applications, there is a surge of research interest in the lifelong learning, especially in the vision domain \citep{rusu2016progressive,kirkpatrick2017overcoming,zenke2017continual,lopez2017gradient,yoon2017lifelong,sprechmann2018memory,chaudhry2018efficient}.
However, its application to language learning has been relatively less studied.
While progress in large-scale unsupervised pretraining \citep{devlin2018bert,radford2019language,liu2019roberta,yang2019xlnet,raffel2020exploring} has recently driven significant advances in the field of natural language processing (NLP), these models require large amounts of in-domain training data and are prone to catastrophic forgetting when trained on new tasks \citep{yogatama2019learning}, hindering their deployment in industry or other realistic setups where new tasks/domains continuously emerge.

One successful approach to achieving lifelong learning has been augmenting the learning model with an episodic memory module \citep{sprechmann2018memory}. 
The underlying idea is to first store previously seen training examples in memory, and later use them to perform experience replay \citep{rolnick2019experience} or to derive optimization constraints \citep{lopez2017gradient,chaudhry2018efficient} while training on new tasks.
Recently, \citet{d2019episodic} propose to use such a memory module for sparse experience replay and local adaptation in the language domain, achieving state-of-the-art results for lifelong learning on text classification and question answering tasks.
Despite its success, the method has three critical downsides, which we demonstrate later in our experiments:
\begin{itemize}
    \item It requires an unrealistically large memory module, i.e. storing \textbf{all} training examples, in order to achieve optimal performance.
    \item While the model can mitigate catastrophic forgetting, its local adaptation step is prone to negative transfer such that it performs worse on the most recent task than the naive baseline without any lifelong learning regularization.
    \item Its inference speed is extremely slow due to a non-trivial amount of local adaptation steps required for \textbf{each} test example.
\end{itemize}

In this paper, we address these limitations and tackle the problem of \textit{efficient} lifelong language learning. 
That is, we focus on storing limited training examples in memory. 
Our contributions are three-fold:
First, we identify three common principles underlying lifelong learning methods. 
We seek to characterize them in language learning and glean insights on overlooked downsides of the existing method.
Second, stemming from this analysis, we propose a meta-lifelong framework that unifies these three principles.
Our approach is a direct extension of \citet{d2019episodic} and it explicitly meta-learns the model as a better initialization for local adaptation.   
Finally, we conduct extensive experiments to demonstrate that our proposed approach can use the identified three principles to achieve efficient lifelong language learning.
We find that our framework outperforms prior methods while using \textit{100 times} less memory storage.
Moreover, we demonstrate that our method can effectively alleviate catastrophic forgetting and negative transfer, closing the performance gap with the multi-task learning upper bound. 
It can also potentially obtain \textit{22 times} faster inference speed.

\section{Background: Principles of Lifelong Language Learning}
\label{sec:background}
Following prior work \citep{d2019episodic}, we consider the lifelong learning setting where a model needs to learn multiple tasks in a sequential order via a stream of training examples without a task descriptor, i.e. the model does not know which task an example comes from during both training and testing.
This setup is ubiquitous in practice, as environments consistently evolve without sending an explicit signal.

Formally, during training, the model makes a single pass over the training example stream consisting of $N$ tasks in an ordered sequence, $\mathcal{D}^{train} = \{\mathcal{D}^{train}_{1}, \cdots, \mathcal{D}^{train}_{N}\}$, where $\mathcal{D}^{train}_{t} = \{(\bm{x}_t^i, y_t^i)\}_{i=1}^{n_t}$ is drawn from the task-specific distribution $P_t(\mathcal{X}, \mathcal{Y})$ of the $t$-th task.
Overall, the goal is to learn a predictor $f_{\theta}: \mathcal{X} \rightarrow \mathcal{Y}$ such as a neural network, parameterized by $\theta \in \mathbb{R}^P$, to minimize the average expected risk of all $N$ tasks:
\begin{equation}
    R(f_{\theta}) \coloneqq \frac{1}{N} \sum_{t=1}^{N} \mathbb{E}_{\bm{x}, y \sim P_t} \left[ \ell(f_{\theta}(\bm{x}), y) \right], 
\label{eq:expected_risk}
\end{equation}
with $\ell$ being the specific task loss.
Notice that while the average risk is most commonly evaluated after the model has seen all tasks, we can also evaluate a specific task at different stages to demonstrate the model's training behavior, and evaluate its robustness against catastrophic forgetting and negative transfer.

While different methods have been developed to optimize Eq.\eqref{eq:expected_risk}, we abstract away from their specific assumptions and instead focus on identifying common principles, among which we stress the following three points that are most relevant to language learning: 

\textbf{Generic Representation.} 
Stemming from transfer learning \citep{weiss2016survey,ganin2015unsupervised}, a key idea of transferring knowledge across diverse tasks is to learn a generic representation (such as a neural network encoder) that is able to encode useful information for all tasks. 
For instance, regularization based lifelong learning methods \citep{kirkpatrick2017overcoming,zenke2017continual,schwarz2018progress,chaudhry2018efficient} add an extra constraint to prevent the model parameter $\theta$ from drastically deviating when training on new tasks, thereby learning a generic model for old tasks as well.
In the language domain, as language models have proven success to generate highly generic representation for many language understanding tasks \citep{yogatama2019learning,raffel2020exploring}, both \citet{d2019episodic} and \citet{sun2020lamal} propose utilizing a pretrained language model \citep{devlin2018bert,radford2019language} to initialize parameters, and further training the model on $\mathcal{D}^{train}$.

\textbf{Experience Rehearsal.}
Motivated by the complementary learning systems (CLS) theory \citep{mcclelland1995there} that humans rely on episodic memory to store past experiences and conduct experience rehearsal, we can also retrain lifelong learning models on previously seen tasks to reduce forgetting.
While prior methods use memory to define optimization constraints \citep{lopez2017gradient,chaudhry2018efficient, sodhani2020toward}, recent work use either stored examples \citep{sprechmann2018memory} or generated synthetic data \citep{shin2017continual,sun2020lamal} to perform experience replay.
Further, \citet{d2019episodic} shows that a sparse 1\% rate of replaying to learning new examples is sufficient for lifelong language learning.

\textbf{Task-specific Finetuning.}
In multi-task learning, injecting task-specific parameters and finetuning on individual task have proven effective for different language understanding tasks \citep{houlsby2019parameter} or even diverse languages \citep{bapna2019simple}.
Prior work \citep{rusu2016progressive,yoon2017lifelong} exploit this idea to expand model parameters for new tasks in lifelong learning setting.
However, all these methods require a task descriptor in order to know when to add new parameters.
When no such signal exists, local adaptation \citep{sprechmann2018memory} uses $K$ stored nearest neighbors of each test example to perform extra finetuning at inference time.
Recent work \citep{d2019episodic,Khandelwal2020Generalization} demonstrate that the sentence embeddings produced by pretrained models can be used to effectively measure query similarity and that local adaptation can improve performance on text classification, question answering and language modelling.

\section{Proposed Framework}

With these principles in mind, we next turn to the problem of how to achieve efficient lifelong learning.
To motivate our proposed framework, we first review the state-of-the-art method, improved MbPA \citep{d2019episodic}, and show how these principles help us to identify the limitation.

\subsection{Model-based Parameter Adaptation}

As a notable example, a recent line of work \citep{sprechmann2018memory,d2019episodic,Khandelwal2020Generalization} have successfully utilized an episodic memory module as a crucial building block for general linguistic reasoning.
Specfically, the improved Model-based Parameter Adaptation (MbPA++) \citep{d2019episodic} consists of three main components: (i) a predictor network $f_{\theta}$, (ii) a key network $g_{\phi}$, and (iii) a memory module $\mathcal{M}$.
The end goal is to train $f_{\theta}$ to generalize well across all tasks as in Eq.\eqref{eq:expected_risk}.

To learn a generic representation, MbPA++ utilizes any state-of-the-art text encoder, such as BERT, to initialize both predictor network $f_{\theta}$ and key network $g_{\phi}$.
At each time step, the model receives a training example $(\bm{x}_t^i, y_t^i) \in \mathcal{D}^{train}$ and updates parameter $\theta$ by optimizing the task loss:
\begin{equation}
    \mathcal{L}_\text{TASK}(\theta; \bm{x}_t^i, y_t^i) = \ell(f_{\theta}(\bm{x}_t^i), y_t^i),
\label{eq:task}
\end{equation}
To determine if the training example should be added to the memory module $\mathcal{M}$, a Bernoulli random variable is drawn with pre-set probability, which is used to control the memory size.

For experience rehearsal, a subset $\mathcal{S}$ of $\mathcal{M}$ is randomly selected, based on a set ratio of replay examples to learning new examples (i.e. revisit $n_{re}$ examples for every $n_{tr}$ training examples).
To avoid catastrophic forgetting, the model then updates the following replay loss to adapt $\theta$ towards seen tasks:
\begin{equation}
    \mathcal{L}_\text{REP}(\theta; \mathcal{S}) = \frac{1}{n_{re}} \sum_{\bm{x}, y \in \mathcal{S}} \ell(f_{\theta}(\bm{x}), y),
\label{eq:replay}
\end{equation}

At inference time, the key network $g_{\phi}$, which is fixed during training, is used to encode example inputs as keys to obtain the $K$ nearest neighbour context $\mathcal{N}_{\bm{x}_i}$ of the $i$-th testing example $\bm{x}_i$.
$L$ local adaptation gradient updates are then performed to achieve task-specific finetuning for the following objective:
\begin{align}
    \mathcal{L}_\text{LA}(\tilde{\theta}_i; \theta, \mathcal{N}_{\bm{x}_i}) &= \frac{1}{K}\sum_{\bm{x}, y \in \mathcal{N}_{\bm{x}_i}} \ell(f_{\tilde{\theta}_i}(\bm{x}), y) \nonumber \\
    & + \lambda_l \|\tilde{\theta}_i-\theta\|_2^2 
\label{eq:local_adaptation}
\end{align}
where $\lambda_l$ is a hyperparameter. The predictor network $f_{\tilde{\theta}_i}$ is then used to output the final prediction for the $i$-th testing example.

Despite its effectiveness, the performance gain of MbPA++ comes at a cost of large memory storage and slow inference speed.
The root of this inefficiency is the non-synergistic nature of the method - the three principles are performed independently without close interaction.
In particular:
(i) the generic representation learned is not optimized for local adaptation and thus more steps are required for robust performance,
(ii) the memory module is selected randomly and lacks a systematic selection method to effectively reduce its size,
(iii) local adaptation only utilize a few neighbours for each testing example so it is prone to overfit and negative transfer when memory size is small.

\subsection{Synergistic Meta-lifelong Framework}

We notice that there is a discrepancy between training and testing in MbPA++.
Specifically, the generic representation is trained on the task loss in Eq.(\ref{eq:task}) directly while it makes prediction \emph{after} the local adaptation at test time. 
Therefore, the model always overfits to the latest task it has seen, and it never learns how to incorporate experience rehearsal efficiently.
According to the CLS theory \cite{mcclelland1995there}, however, human learning systems are complementary in nature - we learn structured knowledge in a manner that allows us to adapt to episodic information fast.
Thus, to resolve the training-testing discrepancy of MbPA++, we change the training goal of generic representation from \emph{how to perform better on the current task} to \emph{how to adapt to episodic memory efficiently}.

\begin{algorithm}[t!]
   \caption{Meta-MbPA}
   \label{algorithm}
\begin{algorithmic}[1]
    \STATE {\bfseries Procedure Train}
    \STATE {\bfseries Input:} training data $\mathcal{D}^{train}$
    \STATE {\bfseries Output:} parameters $\theta$, memory $\mathcal{M}$
    \STATE Initialize $\theta$ with some pretrained model
    \FOR{$(\bm{x}_t^i, y_t^i) \in \mathcal{D}^{train}$}
    \STATE {\bfseries [Generic Representation]} Perform a gradient update on $\theta$ to minimize Eq.(5)
    \IF{training step mod $n_{tr}$ = 0}
    \STATE Sample $n_{re}$ examples from $\mathcal{M}$
    \STATE {\bfseries [Experience Rehearsal]} Perform a gradient update on $\theta$ to minimize Eq.(6)
    \ENDIF
    \STATE Compute $p(\bm{x}_t^i)$ according to Eq.(7)
    \IF{Bernoulli($p(\bm{x}_t^i)$) = 1}
    \STATE Update memory $\mathcal{M} \leftarrow \mathcal{M} \cup (\bm{x}_t^i, y_t^i)$
    \ENDIF
    \ENDFOR
    \STATE {\bfseries Procedure Test}
    \STATE {\bfseries Input:} test examples $\bm{x}$ 
    \STATE {\bfseries Output:} predictions  $\bm{\hat{y}}$
    \FOR{$l=1,...,L$}
    \STATE Sample K examples from $\mathcal{M}$
    \STATE {\bfseries [Task-specific Finetuning]} Perform a gradient update on $\theta$ to minimize Eq.(4)
    \ENDFOR
    \STATE Output prediction $\bm{\hat{y}}_i = f_{\theta}(\bm{x}_i)$
\end{algorithmic}
\end{algorithm}

In particular, we propose an extension of MbPA++ that exploits a meta learning paradigm to interleave the three key principles:
(i) to resolve the training-testing gap, our framework learns a generic representation that is tailored for local adaptation, 
(ii) to enable robust local adaptation, the memory module uses a diversity-based selection criteria to reduce memory size,
(iii) to accommodate small memory, the framework utilizes a coarse local adaptation to alleviate negative transfer.
The full framework is outlined in Algorithm \ref{algorithm} 
and below we detail how each principle is instantiated in a systematic way.

\textbf{Generic Representation.}
We incorporate local adaptation into training generic representation.
In particular, we exploit the idea of meta learning by formulating local adaptation as the base task and representation learning as the meta task.
That is, the generic representation is trained such that it should perform well \emph{after} the local adaptation (a.k.a. learning to adapt).
Thus, for each training example $(\bm{x}_t^i, y_t^i) \in \mathcal{D}^{train}$, we formulate the task loss in Eq.\eqref{eq:task} into a meta-task loss as:
\begin{equation}
\label{eq:meta_task}
\begin{aligned}
\mathcal{L}^{\text{meta}}_\text{TASK}(\theta; \bm{x}_t^i, y_t^i) = \ell(f_{\tilde{\theta}_{\bm{x}_t^i}}(\bm{x}_t^i), y_t^i)\\
\text{s.t.} \quad \tilde{\theta}_{\bm{x}_t^i} = \theta - \alpha \nabla_{\theta} \mathcal{L}_\text{LA}(\theta; \mathcal{N}_{\bm{x}_t^i}) \\
\end{aligned}
\end{equation}
where $\alpha$ is the current learning rate.
Notice the differentiation requires computing the gradient of gradient, which can be implemented by modern automatic differentiation frameworks.
Intuitively, we first approximate local adaptation using gradient step(s), and then optimize the adapted network.

\textbf{Experience Rehearsal.}
With similar rationale to the meta-task loss, we reformulate the memory replay loss in Eq.\eqref{eq:replay} into a meta-replay loss:
\begin{equation}
\label{eq:meta_replay}
\begin{aligned}
\mathcal{L}^{\text{meta}}_\text{REP}(\theta; \mathcal{S}) = \frac{1}{n_{re}} \sum_{\bm{x}, y \in \mathcal{S}} \ell(f_{\tilde{\theta}_{\bm{x}}}(\bm{x}), y) \\
\text{s.t.} \quad \tilde{\theta}_{\bm{x}} = \theta - \alpha \nabla_{\theta} \mathcal{L}_\text{LA}(\theta; \mathcal{N}_{\bm{x}})
\end{aligned}
\end{equation}
with the objective to stimulate efficient local adaptation for all tasks.

We use the same replay ratio as in MbPA++ to keep the meta replay sparse.
In addition, we propose a diversity-based selection criterion to determine if a training example $(\bm{x}_t^i, y_t^i) \in \mathcal{D}^{train}$ should be added to the memory module.
Here, we exploit the key network $g_{\phi}$ to estimate diversity via the minimum distance of $\bm{x}_t^i$ to existing memory as:
\begin{equation}
\label{eq:memory}
\log(p(\bm{x}_t^i)) \propto - \frac{\displaystyle \min_{\bm{x}, y \in \mathcal{M}} \|g_{\phi}(\bm{x}_t^i)-g_{\phi}(\bm{x})\|^2_2}{\beta},
\end{equation}
where $p(\bm{x}_t^i)$ is the probability of the example being selected and $\beta$ is a scaling parameter.
The intuition is to select examples that are less similar to existing memory thereby covering diverse part of data distribution.
As shown later, the proposed method outperforms the uncertainty-based selection rule \citep{ramalho2019adaptive}, which picks examples based on certainty level of the predictor network $f_{\theta}$.
This is because local adaptation is prone to negative transfer when the memory $\mathcal{M}$ misrepresents the true data distribution.

\begin{table*}[t!]
\begin{center}
\resizebox{\textwidth}{!}{
\begin{tabular}{l | c c c c c c c | c c c}
\toprule
    Order & Enc-Dec & Online & A-GEM$^{\dagger}$ & Replay & MbPA++$^{\dagger}$ & MbPA++ & Meta-MbPA & MTL & MTL & LAMOL$^{\ddagger}$ \\
    & & EWC & & & & (Our Impl.) & (1\%) &  & (1\%) &  \\
\bottomrule
\multicolumn{11}{c}{Text Classification}\\
\Xhline{\arrayrulewidth}    
i. & 35.5 & 43.8 & 70.7 & 63.4 & 70.8 & 75.3  & \bf 77.9 & - & - & 76.7 \\
ii. & 44.8 & 49.8 & 65.9 & 73.0 & 70.9 & 74.6  & \bf 76.7 & - & - & 77.2 \\
iii. & 42.4 & 59.5 & 67.5 & 65.8 & 70.2 & 75.6  & \bf 77.3 & - & - & 76.1 \\
iv. & 28.6 & 52.0 & 63.6 & 74.0 & 70.7 & 75.5  & \bf 77.6 & - & - & 76.1 \\
\Xhline{\arrayrulewidth}
Average & 37.8 & 51.3 & 66.9 & 69.1 & 70.6 & 75.3 & \bf 77.3 & 78.9 & 50.4 & 76.5 \\
\bottomrule
\multicolumn{11}{c}{Question Answering}\\
\Xhline{\arrayrulewidth}
i. & 60.9 & 58.0 & 56.1 & 62.3 & 62.0 & 63.3  & \bf 64.8 & - & - & - \\
ii. & 57.3 & 57.2 & 58.4 & 61.3 & 62.4 & 63.5  & \bf 65.3 & - & - & - \\
iii. & 47.0 & 49.5 & 52.4 & 58.3 & 61.4 & 61.6  & \bf 64.4 & - & - & - \\
iv. & 61.0 & 58.7 & 57.9 & 62.9 & 62.4 & 62.4  & \bf 65.0 & - & - & - \\
\Xhline{\arrayrulewidth}
Average & 56.6 & 55.9 & 56.2 & 61.2 & 62.1 & 62.7 & \bf 64.9 & 68.6 & 44.1 & - \\
\bottomrule
\end{tabular}
}
\end{center}
\caption[caption]{\textbf{Accuracy and $F_1$ scores for text classification and question answering, respectively.} Methods that use the defined lifelong learning setup in Section \ref{sec:background} are listed on the left. Where applicable, all methods use $r_{\mathcal{M}}=100\%$ memory size unless denoted otherwise. The best result for lifelong learning methods is made \textbf{bold}. $\dagger$ Results obtained from \citep{d2019episodic}. $\ddagger$ LAMOL \citep{sun2020lamal} is not directly comparable due to their different problem setup where task descriptors are available.}
\vskip -0.1in
\label{tab:summary}
\end{table*}

\textbf{Task-specific Finetuning.}
With small memory, local adaptation for each testing example is prone to negative transfer.
This is because less related memory samples are more likely to be included in $\mathcal{N}_{\bm{x}_i}$ and the model can easily overfit.
Thus, we consider local adaptation with more coarse granularity.
For example, we can cluster testing examples and conduct local adaptation for each cluster independently.
In our experiments, we find that it is sufficient to take this to the extreme such that we consider all test examples as a single cluster.
Consequently, we consider the whole memory as neighbours and we randomly sample from it to be comparable with the original local adaptation formulation (i.e. same batch sizes and gradient steps). 
As shown in the next section, it has two benefits: (1) it is more robust to negative transfer, (2) it is faster when we evaluate testing examples as a group.


\section{Experiments}
\label{sec:exp}

\subsection{Evaluation Dataset}
To evaluate the proposed framework, we conduct experiments on text classification and question answering tasks (see Appendix \ref{sec:dataset} for details). 
Following prior work, we consider each dataset as a separate task and the model needs to sequentially learn several tasks of the same category (e.g. all text classification tasks).
As pointed out in \citep{mccann2018natural}, many NLP tasks can be formulated as question answering and thus our setup is general.

\paragraph{Text classification} We use five datasets from \citep{zhang2015character} spanning four text classification tasks: (1) news classification (AGNews), (2) sentiment analysis (Yelp, Amazon), (3) Wikipedia article classification (DBPedia) and (4) questions and answers categorization (Yahoo). To compare our framework with \cite{d2019episodic}, we follow the same data processing procedure as described by them to produce balanced datasets. In total, we have $33$ classes, $575,000$ training examples and $38,000$ test examples from all datasets. 

\paragraph{Question Answering} Following \cite{d2019episodic}, we use three question answering datasets: SQuAD v1.1\cite{rajpurkar2016squad}, TriviaQA \citep{joshi2017triviaqa} and QuAC \cite{choi2018quac}. TriviaQA has two sections, Web and Wikipedia, which we consider as separate datasets. We process the datasets to follow the same setup as \cite{d2019episodic}. Our processed datasets includes $60,000$-$90,000$ training and $7,000$-$10,000$ validation examples per task.

\subsection{Experimental Setup}

We consider the prominent baselines corresponding to each one of the three principles as introduced in Section \ref{sec:background}. We first consider a standard encoder-decoder model (\textit{Enc-Dec}) which does not utilize any lifelong learning regularization. In the spirit of learning generic representation using parameter regularization, we compare our framework with \textit{Online EWC} \citep{schwarz2018progress} and \textit{A-GEM} \citep{chaudhry2018efficient}. For experience rehearsal, we implement \textit{Replay}, a model that uses stored examples for sparse experience replay only. Finally, we compare with the state-of-the-art \textit{MbPA++} \citep{d2019episodic} which combines experience rehearsal with task-specific finetuning.



\textbf{Implementation Details} We utilize the pre-trained $\text{BERT}_{\text{BASE}}$ \citep{Wolf2019HuggingFacesTS} for initializing the encoder network. $\text{BERT}_{\text{BASE}}$ has $12$ Transformer layers, $12$ self-attention heads, and $768$ hidden dimensions (110M parameters). Similar to \citep{d2019episodic}, we use a separate pre-trained $\text{BERT}_{\text{BASE}}$ for key network and freeze it to prevent from drifting while training on a non-stationary data distribution. For text classification, we use encoded representation of the special beginning-of-document symbol \verb [CLS] as our key. For question answering, we use the question part of the input to get the encoded representation. For both tasks, we store the input example as its associated memory value. Further, we use Faiss \citep{johnson2019billion} for efficient nearest neighbor search in the memory, based upon the key network.

We mainly set hyper-parameters as mentioned in \citep{d2019episodic}. We use Adam \citep{kingma2014adam} as our optimizer, set dropout \citep{srivastava2014dropout} to $0.1$ and the base learning rate to $3e^{-5}$. For text classification, we use a training batch of size $32$ and set the maximum total input sequence length after tokenization to $128$. For question answering, we use a training batch of size $8$, set the maximum total input sequence length after tokenization to $384$ and to deal with longer documents we set document stride to $128$. We also set the maximum question length to $64$.

For Online EWC \citep{schwarz2018progress}, we set the regularization strength $\lambda=5000$ and forgetting coefficient $\gamma=0.95$. For all models with memory module (Replay, MbPA++, Meta-MbPA), we replay $100$ examples for every $10,000$ new examples, i.e., $n_{tr}=10,000$ and
$n_{re}=100$. As mentioned in \citep{d2019episodic}, for MbPA++, we set the number of neighbors $K=32$, the number of local adaptation steps $L=30$ and $\lambda_l=0.001$. We tune the local adaptation learning rate $\alpha$ for MbPA++ in our re-implementation (MbPA++ Our Impl.) and report the improved numbers as well as their reported numbers in Table \ref{tab:summary}, \ref{tab:dataset_specific_results_tc}, and \ref{tab:dataset_specific_results_qa}. For text classification, we set $\alpha=5e^{-5}$ and for question answering we set $\alpha=1e^{-5}$. 

For our framework, Meta-MbPA\footnote{Source code is available at \url{https://github.com/sanketvmehta/efficient-meta-lifelong-learning}.}, unless stated otherwise, we set the number of neighbors $K=32$ and control the memory size through a write rate $r_{\mathcal{M}}=1\%$.
We use $L=30$ local adaptation steps and perform local adaptation for whole testing set.
That is, we randomly draw $K=32$ examples from the memory and perform a local adaptation step.
Through this, the computational cost is equivalent to MbPA++ but we only need to perform the whole process once while MbPA++ requires conducting local adaptation independently for each testing example.
We set  $\alpha=1e^{-5}$ (in Eq. \eqref{eq:meta_task}, \eqref{eq:meta_replay}), $\beta=10$ (in Eq. \eqref{eq:memory}) and $\lambda_l=0.001$ (in Eq. \eqref{eq:local_adaptation}). All of the experiments are performed using PyTorch \citep{paszke2017automatic}, which allows for automatic differentiation through the gradient update as required for optimizing the meta-task loss Eq. \eqref{eq:meta_task} and meta-replay loss Eq. \eqref{eq:meta_replay}.

\subsection{Results}

We use four different orderings of task sequences as in \citep{d2019episodic} (see Appendix \ref{sec:dataset}) and evaluate the model at the end of all tasks.
Following prior work, we report the macro-averaged accuracy for classification and $F_1$ score for question answering. Table \ref{tab:summary} provides a summary of our main results. 
Notice that results on the right are not comparable due to different setups.
The complete per-task results are available in Appendix \ref{sec:dataset_specific_results}.

We first compare our framework (Meta-MbPA) with all baselines.
Even using only 1\% of total training examples as memory, the proposed framework still outperforms existing methods on both text classification and question answering.
Specifically, while regularization-based methods (A-GEM and Online EWC) perform better than the standard Enc-Dec model, their performance vary depending on the task ordering and thus are not robust.
On the other hand, methods that involve local adaptation (MbPA++ and ours) perform consistently better for all orderings.
In particular, our framework improves over MbPA++ while using 100 times less memory, demonstrating the effectiveness of the proposed approach.

We then compare lifelong learning methods to the multitask model MTL, which serves as an upper bound of achievable performance.
As shown in Table \ref{tab:summary}, there is still a non-trivial gap between MbPA++ and MTL, albeit MbPA++ stores all training examples as memory.
Our framework narrows the gap while using smaller memory.

\subsection{Analysis}

\textbf{Memory Capacity.}
In Table \ref{tab:summary}, MbPA++ uses 100\% memory while our framework only uses 1\% memory.
To test memory efficiency, we present results for models using equivalent memory resources in Table \ref{tab:limited_memory}.
The results demonstrate that the performance of MbPA++ degrades significantly as the memory size decreases. 
Consequently, the performance gap between MbPA++ and Meta-MbPA enlarges when they both use equal amount of stored examples, compared to results in Table \ref{tab:summary}.
It is then natural to ask if using memory alone is sufficient to obtain good performance.
We thus compare with MTL trained on subsampled training data, which is equivalent to only performing local adaptation without training the generic representation.
Notice that this variant of MTL is \emph{not} an upper bound as it uses less resources.
Our method significantly outperforms it, showing that the meta generic representation in our method is also crucial to achieve good performance.
These results validate that the proposed framework can utilize the memory module more effectively than existing methods.

We then study the source of improvement of our method.
In particular, we show that prior method is prone to negative transfer.
To see this, we first conduct a case study of memory selection rule.


\begin{table}
\begin{center}
\begin{small}
\begin{tabular}{l c c c c}
\toprule
 & \multicolumn{2}{c}{$r_{\mathcal{M}}=1\%$} & \multicolumn{2}{c}{$r_{\mathcal{M}}=10\%$} \\
\cmidrule(lr){2-3} \cmidrule(lr){4-5}
Model / Task & class. & QA & class. &  QA\\
\bottomrule
MbPA++ & 73.1 & 61.9 & 73.5 & 62.6\\
Meta-MbPA & 77.3 & 64.9 & 78.0 & 65.5\\
\Xhline{\arrayrulewidth} 
MTL & 50.4 & 44.1 & 70.5 & 56.2 \\
\bottomrule
\end{tabular}
\end{small}
\end{center}
\caption[caption]{\textbf{Performance of models using different sizes of memory.}}
\vskip -0.1in
\label{tab:limited_memory}

\end{table}

\begin{figure}%
    \centering
    \subfloat[Random]{\includegraphics[width=0.24\textwidth]{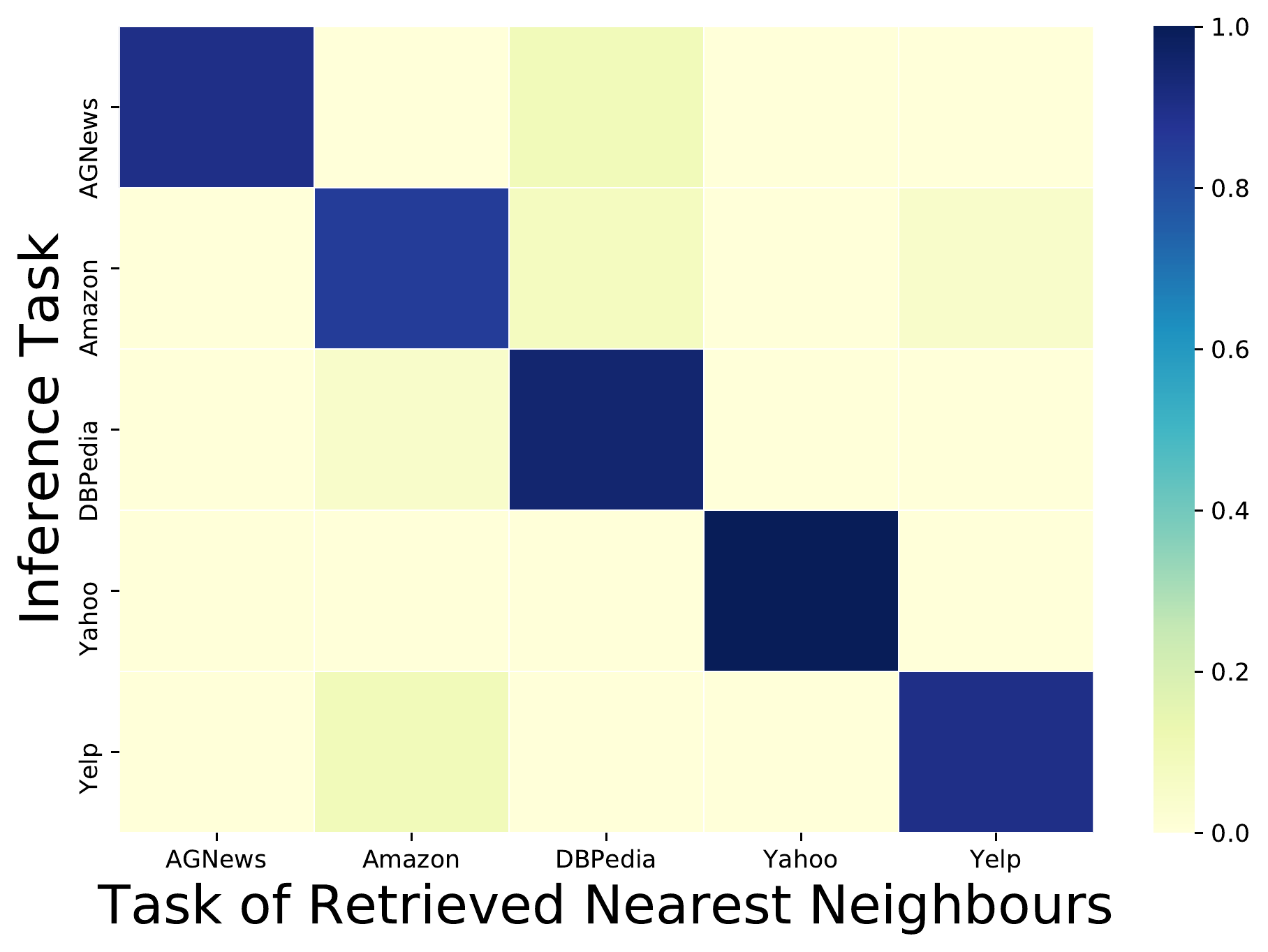} }%
    \subfloat[Uncertainty]{\includegraphics[width=0.24\textwidth]{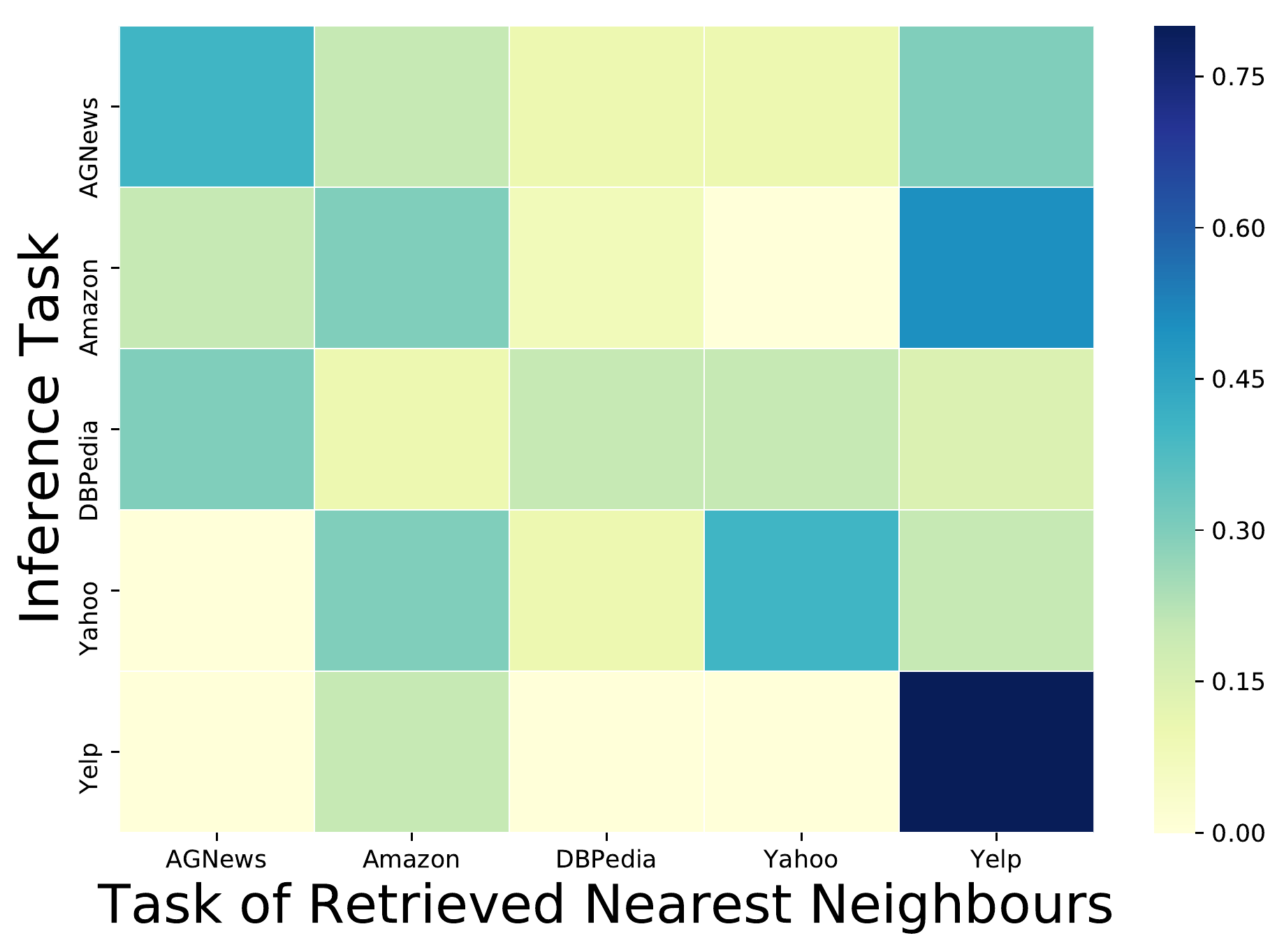} } \\
    \subfloat[Diversity]{\includegraphics[width=0.24\textwidth]{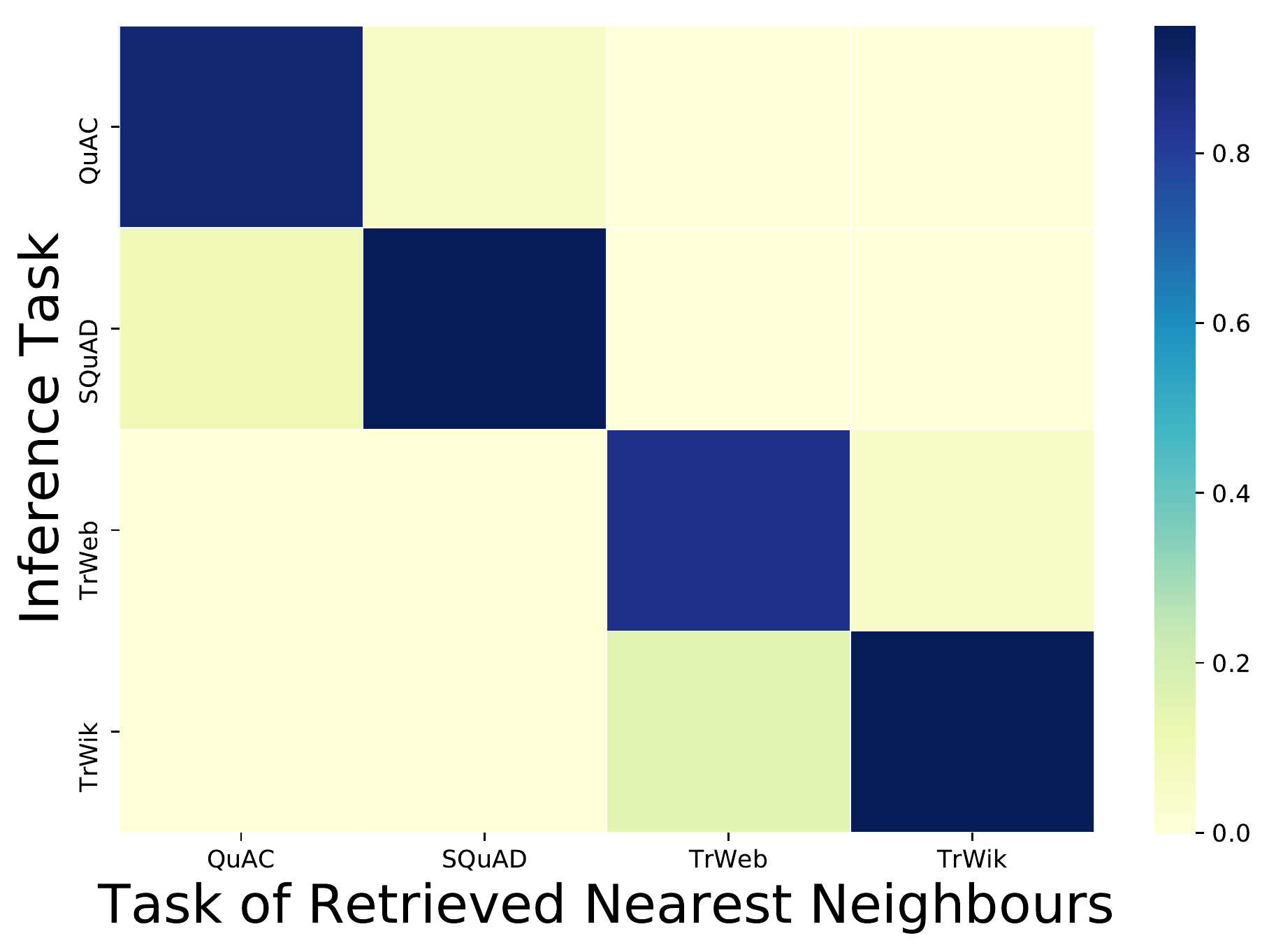} }%
    \subfloat[Forgettable]{\includegraphics[width=0.24\textwidth]{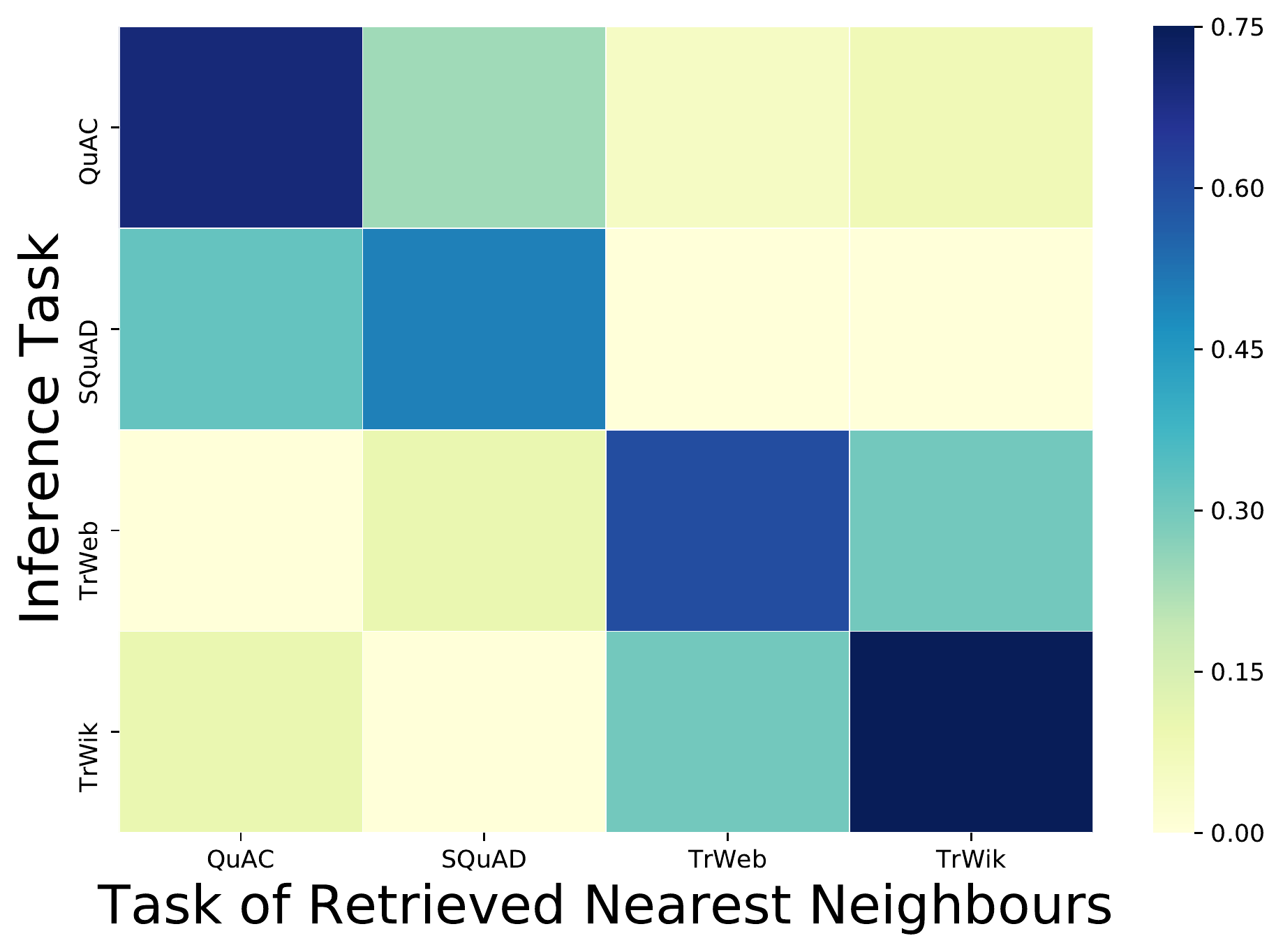} }%
    \caption{\textbf{Proportions of source of neighbours used in local adaptation for each task when different memory selection rule is used, e.g. 10\% of neighbours retrieved for Yelp belong to Amazon.} Numbers in each row sum to 1. Classification figures are at the top while QA at the bottom (Task ordering i.)}%
    \label{fig:memanalysistc}%
\vskip -0.1in
\end{figure}

\textbf{Memory Selection Rule.}
We consider two popular paradigms in active learning \citep{donmez2007dual}, namely the diversity-based method that picks the most representative examples and the uncertainty-based method that picks the most unsure examples.
In particular, we compare four selection criteria belonging to these two categories: random selection, our proposed diversity-based method in Eq.\eqref{eq:memory}, and two uncertainty-based methods \citep{ramalho2019adaptive,toneva2018empirical}.
Notice that random selection is considered as a diversity-based method since it picks examples that represent the true data distribution.
As shown in Table \ref{tab:memory_criteria}, we observe that the choice of memory selection criteria clearly impacts performance.
While the proposed diversity method slightly outperforms random selection, the two uncertain-based methods perform \textit{worse} than the random baseline, consistent with similar findings reported in \citet{d2019episodic}.

\begin{table}[t!]
\begin{center}
\begin{small}
\begin{tabular}{l | c c c}
\toprule
 & Replay & MbPA++ & Meta-MbPA\\
\bottomrule
\multicolumn{4}{c}{Text Classification}\\
\Xhline{\arrayrulewidth}  
Random & 69.2 & 73.1 & 76.8 \\
Diversity & 69.1 & 73.0 & 77.3 \\
Uncertainty & 65.4 & 41.2 & 62.7 \\
Forgettable & 62.7 & 50.5 & 61.8 \\
\bottomrule
\multicolumn{4}{c}{Question Answering}\\
\Xhline{\arrayrulewidth}  
Random & 61.2 & 61.9 & 63.8 \\
Diversity & 61.5 & 62.2 & 64.9 \\
Uncertainty & 56.1 & 50.4 & 54.2\\
Forgettable & 59.7 & 52.1 & 57.5\\
\bottomrule
\end{tabular}
\end{small}
\end{center}
\caption[caption]{\textbf{Performance of models using different memory selection criteria.} ``Uncertainty'' utilizes model's confidence level \citep{ramalho2019adaptive}. ``Forgettable'' picks examples according to forgetting events \citep{toneva2018empirical}. We tune hyperparameters that result in $r_{\mathcal{M}}=1\%$ memory size for all methods.}
\vskip -0.1in
\label{tab:memory_criteria}
\end{table}

We seek an explanation for this phenomenon and visualize the heat maps in Figure \ref{fig:memanalysistc} to show which tasks each testing example's retrieved neighbours come from during the local adaptation phase.
Ideally, the model should always use neighbours from the same task and the heat map should be diagonal.
We observe that, compared to diversity-based methods, more examples from other tasks are used as nearest neighbours when models use uncertainty-based methods.
This is because the selected uncertain examples are usually less representative in the true distribution and could be outliers.
Thus, the resulting memory does not have a good coverage of the data distribution and no similar examples exist for certain test examples.
Consequently, less related examples from other tasks are used for the local adaptation, which causes negative transfer.
This is verified in Table \ref{tab:uncertainty}, where models without local adaptation outperform their locally adapted counterparts.
More importantly, Meta-MbPA obtains much smaller performance gaps, indicating that it is more robust to negative transfer.
We further verify this in the following section.

\begin{table}[t!]
\begin{center}
\begin{small}
\begin{tabular}{l c c c c}
\toprule
 & \multicolumn{2}{c}{Uncertainty} & \multicolumn{2}{c}{Forgettable} \\
\cmidrule(lr){2-3} \cmidrule(lr){4-5}
Model / Task & class. & QA & class. & QA\\
\bottomrule
Meta-MbPA & 62.7 & 54.2 & 61.8 & 57.5\\
\hspace{2mm} w/o LA & 65.8 & 55.8 & 67.9 & 59.2\\
\Xhline{\arrayrulewidth} 
MbPA++ & 41.2 & 50.4 & 50.5 & 52.1 \\
\hspace{2mm} w/o LA & 65.4 & 56.1 & 68.4 & 59.2\\
\bottomrule
\end{tabular}
\end{small}
\end{center}
\caption[caption]{\textbf{Performance of models using the uncertainty-based memory selection methods (correspond to Table \ref{tab:memory_criteria}).} ``LA'' refers to local adaptation.}
\vskip -0.1in
\label{tab:uncertainty}
\end{table}

\begin{figure*}%
    \centering
    \subfloat[Text Classification-AGNews]{\includegraphics[width=0.28\textwidth]{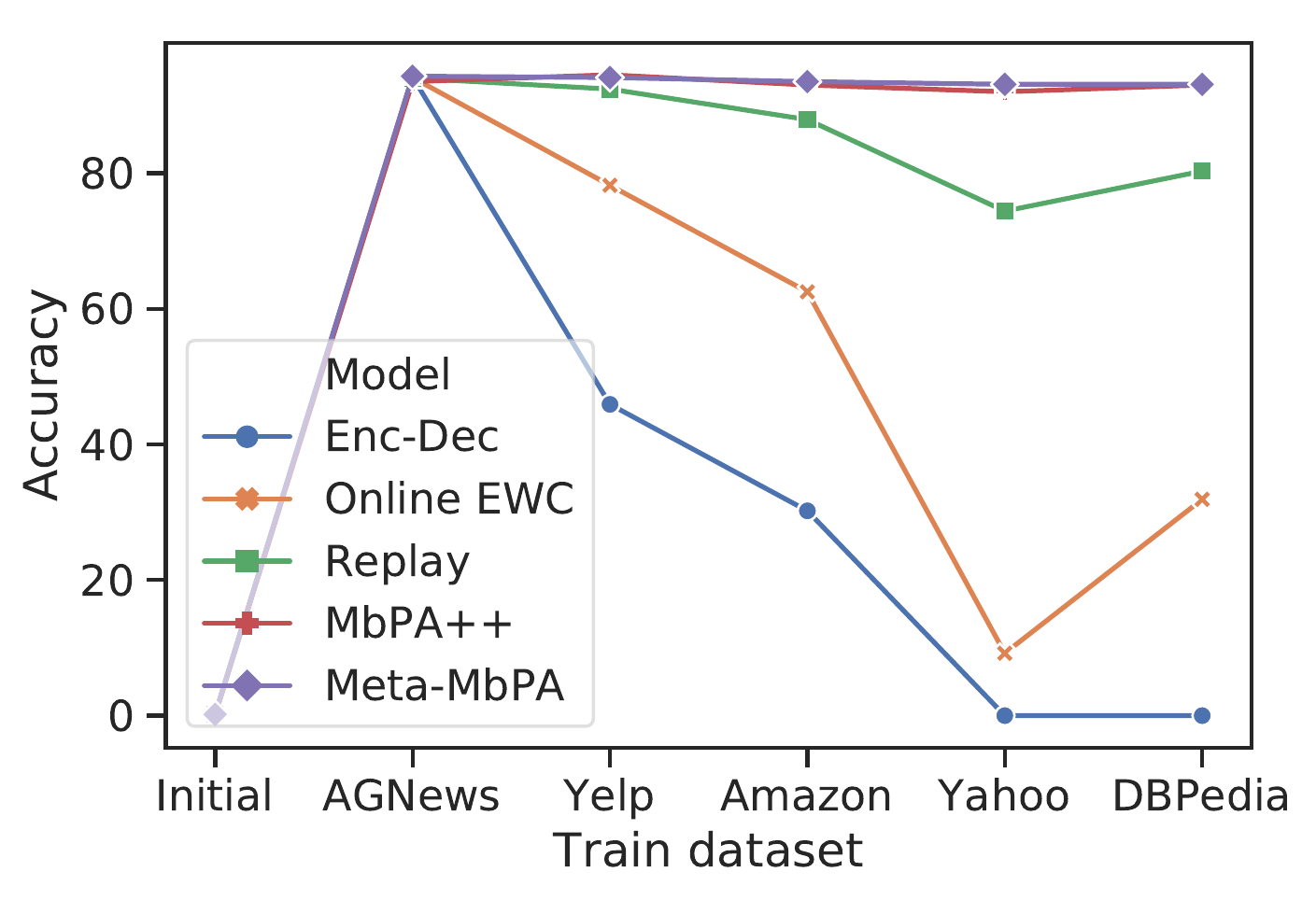} }%
    \qquad
    \subfloat[Question Answering-QuAC]{\includegraphics[width=0.28\textwidth]{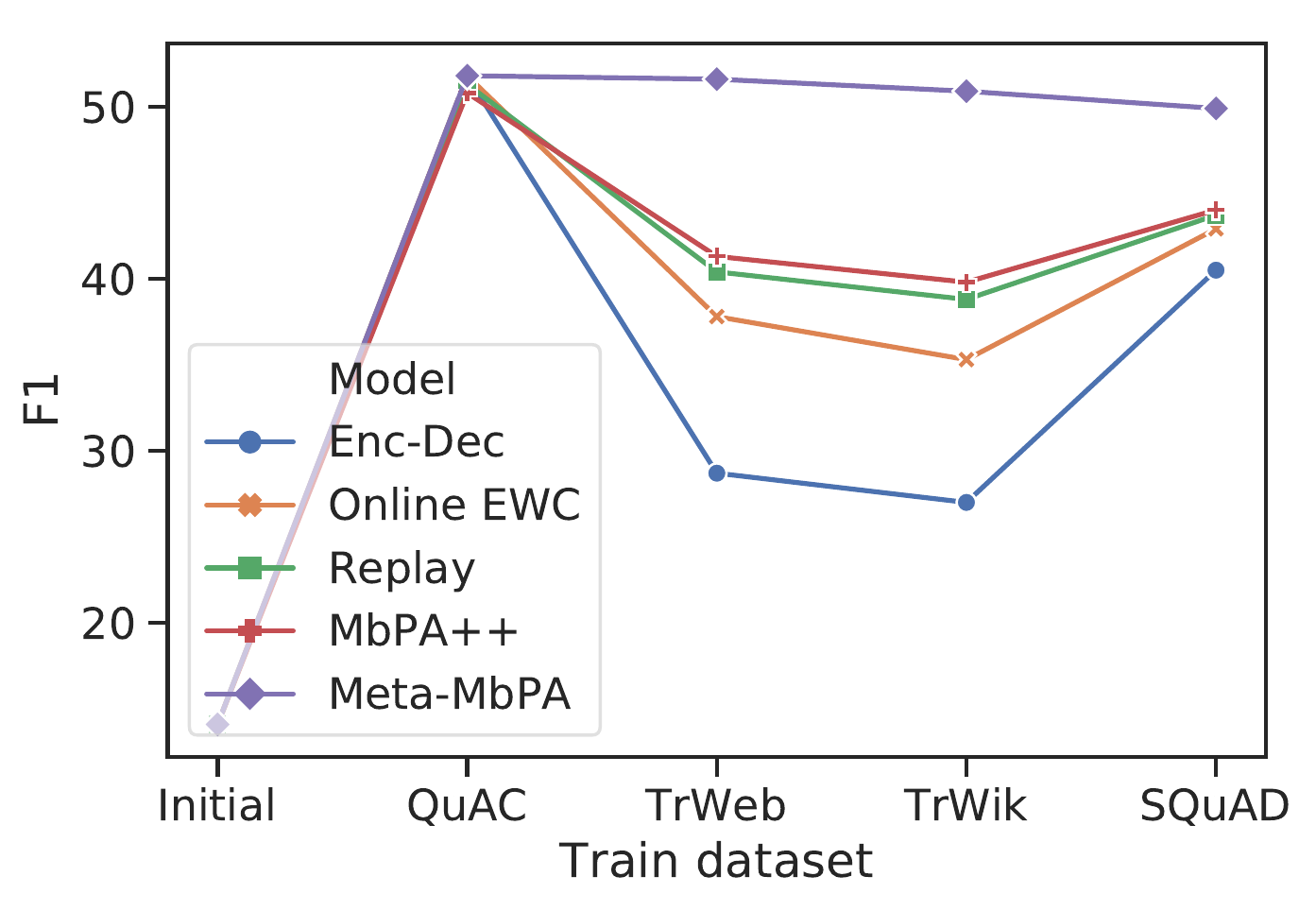} }%
    \qquad
    \subfloat[Question Answering-SQuAD]{\includegraphics[width=0.28\textwidth]{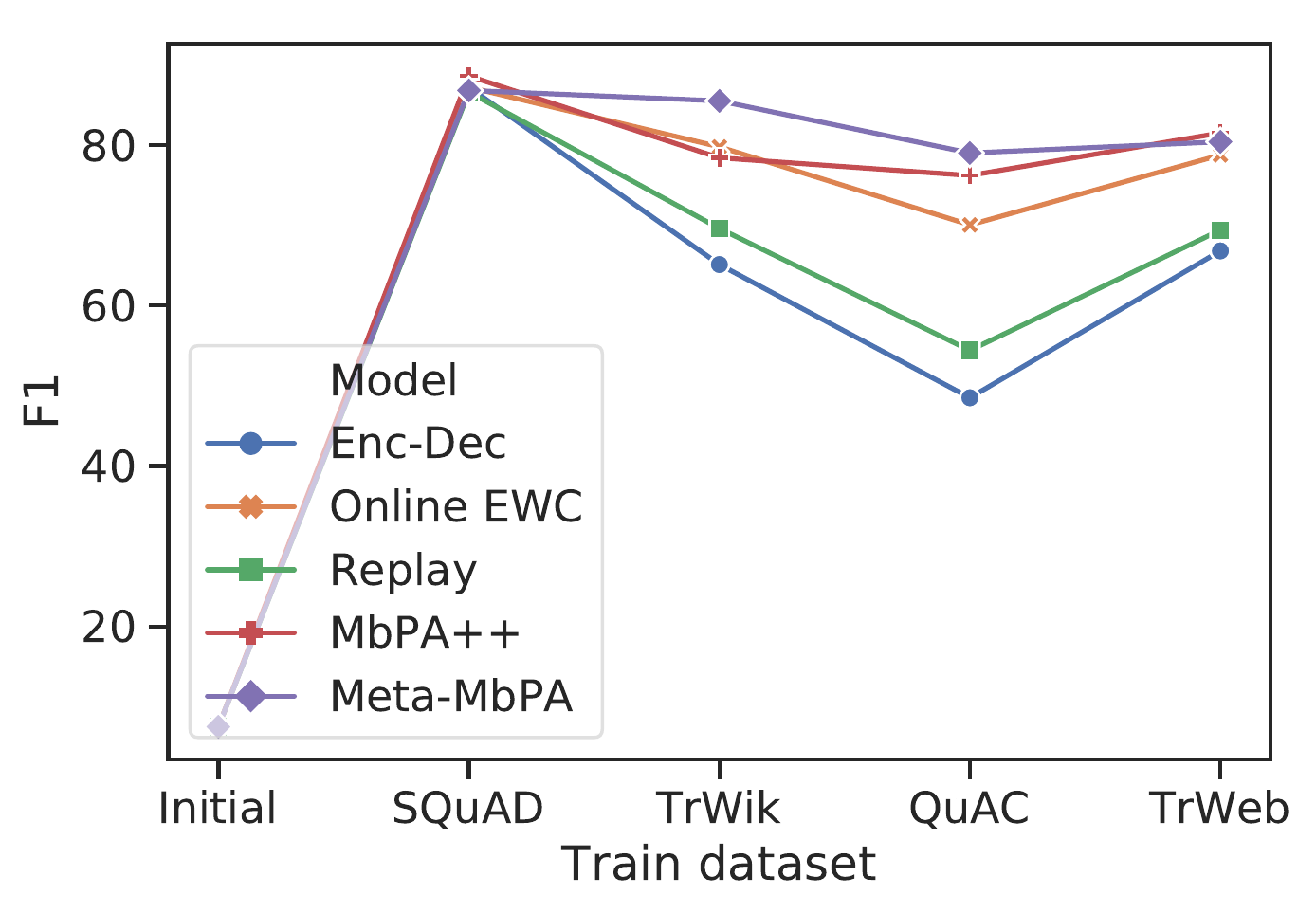} }%
    \caption{\textbf{Catastrophic Forgetting of the first dataset as training progresses.} Complete results in Appendix \ref{sec:first_task_progress}}
    \label{fig:catforget}%
\end{figure*}

\textbf{Trade-off between Catastrophic Forgetting and Negative Transfer.}
We first verify the models' robustness to catastrophic forgetting.
As shown in Table \ref{tab:dataset_specific_results_tc} and \ref{tab:dataset_specific_results_qa} (Appendix \ref{sec:dataset_specific_results}), the standard Enc-Dec model performs poorly on previously trained tasks, indicating the occurrence of catastrophic forgetting.  
While all baselines can alleviate the forgetting to some degree, our framework achieves the best performance on previously learned tasks.
We also evaluate the model's performance on the first task as it continues to learn more tasks.
Figure \ref{fig:catforget} illustrates how each model retains its previously acquired knowledge as it learns new knowledge.
We observe that our framework is consistently better than the baselines at mitigating forgetting.

In addition, as prior work have shown transferring from diversely related sources can hurt performance in the target \citep{ge2014handling,wang2018towards},
we study if transferring from multiple tasks learned in the past can induce negative transfer, which is often overlooked in existing studies on lifelong learning.
Table \ref{tab:negative_transfer} shows the averaged results on the last task in each task ordering (see Appendix \ref{sec:dataset_specific_results} for complete results).
Surprisingly, compared to the Enc-Dec baseline, MbPA++ actually performs \textit{worse} on the last task despite its improved macro-averaged performance (Table \ref{tab:summary}).
This suggests that while it is robust to catastrophic forgetting, MbPA++ fails to utilize prior knowledge to benefit later tasks and thus is prone to negative transfer.
Apart from some practical bottlenecks such as limited model capacity, local adaptation is a critical factor of negative transfer as Replay\footnote{Replay is equivalent to MbPA++ without local adaptation.} outperforms MbPA++ in Table \ref{tab:negative_transfer}.
Intuitively, this shows that since Replay already performs well on the last task, further using local adaption can overfit and hurt the performance.
On the other hand, the proposed method is trained to learn a more robust initialization for adaptation and uses a coarse adaptation that is less prone to negative transfer. 
Therefore, it outperforms MbPA++ and closes the gap with Enc-Dec on the last task, consistent with results in Table \ref{tab:uncertainty}.
All of these experiments illustrate that there is a trade-off between catastrophic forgetting and negative transfer, such that more adaptations are desired for earlier tasks while less is better for later tasks.
While prior studies focus on catastrophic forgetting only, we are the first to show the importance of balancing the trade-off to avoid both negative effects.


\begin{table}[t!]
\begin{center}
\begin{small}
\begin{tabular}{l c c c c}
\toprule
& Enc-Dec & Replay & MbPA++ & Meta-MbPA\\
\bottomrule
class. & 82.1 & 81.8 & 78.6 & 82.1\\
QA & 72.6 & 72.7 & 70.7 & 72.1 \\
\bottomrule
\end{tabular}
\end{small}
\end{center}
\caption[caption]{\textbf{Average performance on the last task across all four task orderings.}}
\vskip -0.1in
\label{tab:negative_transfer}
\end{table}

\begin{table}
\begin{center}
\begin{small}
\begin{tabular}{l c c c c}
\toprule
 & \multicolumn{2}{c}{$r_{\mathcal{M}}=1\%$} & \multicolumn{2}{c}{$r_{\mathcal{M}}=50\%$} \\
\cmidrule(lr){2-3} \cmidrule(lr){4-5}
Model / Task & class. & QA & class. &  QA\\
\bottomrule
Meta-MbPA & 77.3 & 64.9 & 78.2 & 66.1\\
\hspace{2mm} w/o Meta & 73.1 & 58.5 & 74.0 & 59.6\\
\hspace{2mm} w/o MS & 76.8 & 63.8 & 78.1 & 66.1 \\
\hspace{2mm} w/o LA & 75.9 & 62.0 & 75.8 & 62.1\\
\bottomrule
\end{tabular}
\end{small}
\end{center}
\caption[caption]{\textbf{Ablation Study on different memory size.} ``Meta'' refers to the proposed meta optimization in Eq.\eqref{eq:meta_task} and \eqref{eq:meta_replay}.``MS'' denotes memory selection based on Eq.\eqref{eq:memory}. ``LA'' refers to local adaptation.}
\label{tab:ablation}
\vskip -0.3in
\end{table}

\textbf{Ablation Study.}
We report the results of ablation study in Table \ref{tab:ablation} and analyze the effects of the three components in our framework subject to different memory sizes.
First, we observe that the model without the meta learning optimization performs the worst, which shows the importance of learning a generic representation tailored for local adaptation.
More importantly, Meta-MbPA achieves worse performance without any local adaptation step.
This demonstrates that learning the generic representation alone is not sufficient enough, and that the meta learning mechanism and local adaptation are complementary, which mimic the complementary human learning systems in the CLS theory.
Finally, while the diversity-based memory selection rule contributes to the performance gain when we use a small memory module, it becomes less effective as the memory size increases.
This is expected since the memory distribution can well represent the true data distribution with a larger capacity, and thus it demonstrates that the proposed methods mostly contribute to robustly reducing the memory sizes for better efficiency.
Overall, these results validate the effectiveness of each component and highlight the importance of complementary lifelong learning systems.
To the best of our knowledge, this is the first work to formulate the slow learning of structured knowledge as meta task and the fast learning from episodic memory as base task.

\textbf{Inference Speed.} 
The ordinary local adaptation requires customized gradient updates for each testing example and thus it is notoriously slow.
Using 1 Nvidia Tesla V100 GPU and 128 GB of RAM, it takes 66.6 hours and 89.3 hours to evaluate on test classification and question answering, respectively.
On the other hand, we use coarse local adaptation in our method which uses the same updates for all testing examples.
Consequently, it takes 2.9 hours and 4.2 hours for our method to finish the evaluation process, achieving a maximum 22 times speedup.
Notice that in a pure online learning setup, our method will obtain similar inference speed as MbPA++.
In addition, we hypothesize that using a different granularity (e.g. clustering testing examples) is beneficial for tasks that are more conflicting in nature, 
as it can balance the trade-off between overfitting to nearest neighbours of small memory and performing more sample-specific adaptation for each test example.
We leave this exploration for future work.


\section{Conclusion}
In this work, we identify three principles underlying different lifelong language learning methods and show how to unify them in a meta-lifelong framework.
Our experiments demonstrate the effectiveness of the proposed framework on text classification and question answering tasks. 
We report new state-of-the-art results while using $100$ times less memory space. 
These results illustrate that it is possible to achieve efficient lifelong learning by establishing complementary learning systems.
Our analysis also shows that negative transfer is an overlooked factor that could cause sub-optimal performance, and we highlight the importance of balancing the trade-off tween catastrophic forgetting and negative transfer for future work.

\section*{Acknowledgments}
We would like to thank Qizhe Xie, Ruochen Xu, Shagun Sodhani, Yiheng Zhou and  Yulia Tsvetkov for insightful discussions. 
We also thank anonymous reviewers for their valueable feedbacks.

\bibliography{emnlp2020}
\bibliographystyle{acl_natbib}

\clearpage
\section*{Appendix}
\appendix



\section{Dataset and Ordering}
\label{sec:dataset}
\paragraph{Text classification}
We use the following text classification dataset orders for comparing our results with \citep{d2019episodic}: \\
    i. Yelp$\rightarrow$AGNews$\rightarrow$DBPedia$\rightarrow$Amazon$\rightarrow$Yahoo \\
    ii. DBPedia$\rightarrow$Yahoo$\rightarrow$AGNews$\rightarrow$Amazon$\rightarrow$Yelp \\
    iii. Yelp$\rightarrow$Yahoo$\rightarrow$Amazon$\rightarrow$DBpedia$\rightarrow$AGNews \\
    iv. AGNews$\rightarrow$Yelp$\rightarrow$Amazon$\rightarrow$Yahoo$\rightarrow$DBpedia \\

\paragraph{Question Answering} Our processed dataset includes SQuAD with $90,000$ training and $10,000$ validation examples, TriviaQA (Web) with $76,000$ training and $10,000$ validation examples, TriviaQA (Wikipedia) with $60,000$ training and $8,000$ validation examples and QuAC with $80,000$ training and $7,000$ validation examples. We consider following dataset orders for question answering: \\
i. QuAC$\rightarrow$TrWeb$\rightarrow$TrWik$\rightarrow$SQuAD \\
ii. SQuAD$\rightarrow$TrWik$\rightarrow$QuAC$\rightarrow$TrWeb \\
iii. TrWeb$\rightarrow$TrWik$\rightarrow$SQuAD$\rightarrow$QuAC \\
iv. TrWik$\rightarrow$QuAC$\rightarrow$TrWeb$\rightarrow$SQuAD

\section{Dataset Specific Results}
\label{sec:dataset_specific_results}
We report per-dataset specific results for text classification in Table \ref{tab:dataset_specific_results_tc} and for question answering in Table \ref{tab:dataset_specific_results_qa}. For A-GEM and MbPA++ baselines, we obtain results from \citep{d2019episodic}. A-GEM, Replay, MbPA++ and MbPA++ (Our Impl.) methods use $r_{\mathcal{M}}=100\%$ memory size while our proposed framework, Meta-MbPA, and MbPA++(1\%) use $r_{\mathcal{M}}=1\%$ memory size. 
\begin{table*}[h]
\begin{center}
\resizebox{\textwidth}{!}{%
\begin{tabular}{c c|c c c c c c c c}
\toprule
    Order & Dataset & Enc-Dec & Online & A-GEM$^{\dagger}$ & Replay & MbPA++$^{\dagger}$ & MbPA++ & MbPA++ & Meta-MbPA \\
    & & & EWC & & &  & (Our Impl.) & ($r_{\mathcal{M}}=1\%$) & ($r_{\mathcal{M}}=1\%$) \\
\bottomrule
\multirow{6}{*}{i}  & 1 & 2.0 & 29.7 & 42.5 & 49.2 & 45.7 & 59.2 & 54.2 & 62.1\\
 & 2 & 4.3 & 0.1 & 89.8 & 50.1 & 91.6 & 94.0 & 91.0 & 93.7\\
 & 3 & 95.8 & 97.5 & 96.0 & 98.7 & 96.3 & 98.5 & 98.5 & 99.1\\
 & 4 & 1.3 & 18.5 & 56.8 & 45.2 & 54.6 & 57.7 & 56.7 & 60.7\\
 & 5 & 74.2 & 73.2 & 68.2 & 74.0 & 65.6 & 67.2 & 66.7 & 73.8\\
\cline{2-10}
 & Average & 35.5 & 43.8 & 70.7 & 63.4 & 70.8 & 75.3 & 73.4 & 77.9\\
\bottomrule
\multirow{6}{*}{ii}  & 1 & 62.2 & 89.9 & 80.1 & 98.7 & 95.8 & 98.5 & 98.0 & 99.0\\
 & 2 & 0.0 & 0.1 & 50.3 & 54.6 & 63.1 & 69.7 & 61.7 & 70.2\\
 & 3 & 39.4 & 40.3 & 91.3 & 89.3 & 92.2 & 95.0 & 93.0 & 92.5\\
 & 4 & 61.3 & 60.0 & 57.3 & 61.5 & 55.7 & 55.2 & 55.2 & 60.1\\
 & 5 & 61.2 & 58.5 & 50.6 & 61.1 & 47.7 & 54.7 & 52.7 & 61.5\\
\cline{2-10}
 & Average & 44.8 & 49.8 & 65.9 & 73.0 & 70.9 & 74.6 & 72.1 & 76.7\\
\bottomrule
\multirow{6}{*}{iii}  & 1 & 11.4 & 52.5 & 41.1 & 54.8 & 44.3 & 59.2 & 53.7 & 59.6\\
 & 2 & 2.1 & 14.9 & 55.0 & 31.9 & 62.7 & 67.7 & 60.2 & 70.2\\
 & 3 & 12.8 & 40.3 & 54.6 & 52.0 & 54.4 & 58.2 & 60.7 & 63.8\\
 & 4 & 92.5 & 98.0 & 93.3 & 97.4 & 96.2 & 98.5 & 98.0 & 98.9\\
 & 5 & 93.3 & 91.8 & 93.6 & 93.1 & 93.4 & 94.5 & 92.5 & 94.1\\
\cline{2-10}
 & Average & 42.4 & 59.5 & 67.5 & 65.8 & 70.2 & 75.6 & 73.0 & 77.3\\
\bottomrule
\multirow{6}{*}{iv}  & 1 & 0.0 & 31.9 & 90.8 & 80.3 & 91.8 & 94.0 & 91.0 & 93.1 \\
 & 2 & 8.3 & 33.3 & 44.9 & 59.3 & 44.9 & 57.2 & 54.2 & 60.8\\
 & 3 & 3.6 & 22.2 & 60.2 & 59.6 & 55.7 & 59.7 & 61.2 & 61.6\\
 & 4 & 31.8 & 73.5 & 65.4 & 71.9 & 65.3 & 68.7 & 63.7 & 73.6\\
 & 5 & 99.1 & 98.9 & 56.9 & 99.1 & 95.8 & 98.0 & 98.5 & 99.1\\
\cline{2-10}
 & Average & 28.6 & 52.0 & 63.6 & 74.0 & 70.7 & 75.5 & 73.7 & 77.6 \\
\bottomrule
\end{tabular}
}
\end{center}
\caption[caption]{\textbf{Dataset specific accuracy for text classification tasks for different dataset orders and models.} $\dagger$ Results obtained from \citep{d2019episodic}. Where applicable, we use $r_{\mathcal{M}}=100\%$ unless denoted otherwise.}
\vskip -0.1in
\label{tab:dataset_specific_results_tc}
\end{table*}

\begin{table*}[h]
\begin{center}
\resizebox{\textwidth}{!}{%
\begin{tabular}{c c|c c c c c c c c}
\toprule
    Order & Dataset & Enc-Dec & Online & A-GEM$^{\dagger}$ & Replay & MbPA++$^{\dagger}$ & MbPA++ & MbPA++ & Meta-MbPA \\
    & & & EWC & & &  & (Our Impl.) & ($r_{\mathcal{M}}=1\%$) & ($r_{\mathcal{M}}=1\%$) \\
\bottomrule
\multirow{6}{*}{i} & 1 & 40.5 & 42.9 & 36.7 & 44.1 & 47.2 & 44.3 & 42.6 & 49.9\\
 & 2 & 60.1 & 57.4 & 51.8 & 60.7 & 57.7 & 62.9 & 60.0 & 63.1\\
 & 3 & 58.2 & 53.8 & 53.4 & 58.7 & 58.9 & 61.2 & 58.8 & 61.5\\
 & 4 & 85.0 & 77.7 & 82.5 & 85.5 & 84.3 & 84.7 & 86.8 & 84.7\\
\cline{2-10}
 & Average & 60.9 & 58.0 & 56.1 & 62.3 & 62.0 & 63.3 & 62.0 & 64.8\\
\bottomrule
\multirow{6}{*}{ii} & 1 & 66.8 & 78.8 & 64.2 & 73.1 & 72.6 & 80.4 & 81.8 & 80.4\\
 & 2 & 64.2 & 59.5 & 62.5 & 64.2 & 63.4 & 65.3 & 60.7 & 61.5\\
 & 3 & 31.4 & 28.6 & 43.4 & 41.0 & 50.5 & 42.0 & 41.6 & 52.1\\
 & 4 & 66.7 & 61.9 & 63.5 & 66.8 & 63.0 & 66.1 & 64.3 & 67.0\\
\cline{2-10}
 & Average & 57.3 & 57.2 & 58.4 & 61.3 & 62.4 & 63.5 & 62.1 & 65.3\\
\bottomrule
\multirow{6}{*}{iii}  & 1 & 41.6 & 57.2 & 47.6 & 58.7 & 56.0 & 62.0 & 59.4 & 65.7 \\
 & 2 & 38.8 & 51.9 & 47.0 & 54.2 & 56.8 & 53.4 & 57.3 & 59.2 \\
 & 3 & 54.4 & 63.1 & 57.4 & 67.7 & 78.0 & 81.8 & 83.9 & 80.7 \\
 & 4 & 53.1 & 25.5 & 57.4 & 52.7 & 54.9 & 49.0 & 46.9 & 52.1\\
\cline{2-10}
 & Average & 47.0 & 49.5 & 52.4 & 58.3 & 61.4 & 61.6 & 61.8 & 64.4 \\
\bottomrule
\multirow{6}{*}{iv}  & 1 & 58.1 & 60.5 & 54.8 & 59.4 & 59.0 & 58.9 & 60.8 & 61.3 \\
 & 2 & 39.8 & 36.3 & 38.8 & 45.0 & 48.7 & 43.5 & 39.2 & 50.4 \\
 & 3 & 60.5 & 60.4 & 53.4 & 61.6 & 58.1 & 64.2 & 61.3 & 63.7\\
 & 4 & 85.6 & 77.3 & 84.7 & 85.6 & 83.6 & 82.8 & 85.3 & 84.5\\
\cline{2-10}
 & Average & 61.0 & 58.7 & 57.9 & 62.9 & 62.4 & 62.4 & 61.6 & 65.0\\
\bottomrule
\end{tabular}
}
\end{center}
\caption[caption]{\textbf{Dataset specific $F_1$ scores for question answering tasks for different dataset orders and models.} $\dagger$ Results obtained from \citep{d2019episodic}. Where applicable, we use $r_{\mathcal{M}}=100\%$ unless denoted otherwise.}
\vskip -0.1in
\label{tab:dataset_specific_results_qa}
\end{table*}

\section{Single Task and Multi-task Models Results}
We report results for single-task models that uses only single-task data and multi-task learning models using different amounts of training data in Table \ref{tab:single_model_mtl}. For text classification, we report accuracy scores and for question answering, we report $F_1$ scores.

\begin{table*}[h]
\begin{center}
\resizebox{0.85\textwidth}{!}{%
\begin{tabular}{c |c c c c}
\toprule
    Dataset & Single Model & MTL (1\%) & MTL (10\%) & MTL (100 \%) \\
\bottomrule
\multicolumn{5}{c}{Text Classification}\\
\Xhline{\arrayrulewidth}    
    AGNews & 93.6 & 83.1 & 88.7 & 94.0\\
    Amazon & 61.8 & 38.6 & 54.2 & 63.5\\
    DBPedia & 99.2 & 78.1 & 91.4 & 99.3\\
    Yahoo & 74.9 & 15.8 & 65.6 & 75.3\\
    Yelp & 61.9 & 36.4 & 52.8 & 62.6\\
    \cline{1-5}
    Average & 78.28 & 50.4 & 70.5 & 78.9 \\
 \bottomrule
 \multicolumn{5}{c}{Question Answering}\\
 \Xhline{\arrayrulewidth}    
    QuAC & 54.0 & 20.9 & 30.9 & 53.5\\
    SQuAD & 87.8 & 60.5 & 75.2 & 88.1 \\
    Trivia Web & 65.8 & 49.2 & 62.2 & 67.7\\
    Trivia Wikipedia & 62.9 & 45.9 & 56.5 & 64.9\\
    \cline{1-5}
    Average & 67.6 & 44.1 & 56.2 & 68.6 \\
 \bottomrule
\end{tabular}
}
\end{center}
\caption[caption]{\textbf{Single model and Multi-Task Learning (MTL) results for text classification and question answering tasks.} MTL ($X\%$) denotes $X\%$ of the training examples are used per dataset to train MTL models.}
\vskip -0.1in
\label{tab:single_model_mtl}
\end{table*}

\begin{table*}[h]
\begin{center}
\resizebox{0.85\textwidth}{!}{%
\begin{tabular}{c l|c c c c c c}
\toprule
    First & Dataset & Enc-Dec & Online & Replay & MbPA++ & Meta-MbPA \\
    Dataset & & & EWC & & (Our Impl.) & ($r_{\mathcal{M}}=1\%$) \\
\bottomrule
\multicolumn{7}{c}{Text Classification}\\
\Xhline{\arrayrulewidth}
\multirow{6}{*}{AGNews} & 0 (Initial) & 0.2 & 0.2 & 0.2 & 0.2 & 0.2\\
 & 1 (AGNews) & 94.2 & 94.1 & 94.0 & 93.5 & 94.3\\
 & 2 (Yelp) & 45.9 & 78.2 & 92.4 & 94.5 & 94.1\\
 & 3 (Amazon) & 30.2 & 62.5 & 87.9 & 93.0& 93.5\\
 & 4 (Yahoo) & 0.0 & 9.2 & 74.4 & 92.0 & 93.1\\
 & 5 (DBPedia) & 0.0 & 31.9 & 80.3 & 93.0 & 93.1\\
\bottomrule
\multirow{6}{*}{Yelp} & 0 (Initial) & 0.2 & 0.2 & 0.2 & 0.2 & 0.2\\
 & 1 (Yelp) & 62.5 & 62.0 & 62.5 & 57.7 & 62.5\\
 & 2 (Yahoo) & 4.3 & 32.3 & 58.1 & 56.7 & 61.0\\
 & 3 (Amazon) & 60.4 & 61.7 & 60.1 & 55.7 & 61.2\\
 & 4 (DBPedia) & 48.6 & 61.4 & 60.3 & 58.2 & 61.4\\
 & 5 (AGNews) & 11.4 & 52.4 & 54.8 & 57.7 & 59.6\\
\bottomrule
\multicolumn{7}{c}{Question Answering}\\
\Xhline{\arrayrulewidth}
\multirow{5}{*}{QuAC} & 0 (Initial) & 14.1 & 14.1 & 14.1 & 14.1 & 14.1\\
 & 1 (QuAC) & 51.8 & 51.8 & 51.3 & 50.8 & 51.8\\
 & 2 (TrWeb) & 28.7 & 37.8 & 40.4 & 41.3 & 51.6\\
 & 3 (TrWik) & 27.0 & 35.3 & 38.8 & 39.8 & 50.9\\
 & 4 (SQuAD) & 40.5 & 42.9 & 43.7 & 44.0 & 49.9\\
\bottomrule
\multirow{5}{*}{SQuAD} & 0 (Initial) & 7.5 & 7.5 & 7.5 & 7.5 & 7.5\\
 & 1 (SQuAD) & 87.2 & 87.2 & 86.6 & 88.6 & 86.8\\
 & 2 (TrWik) & 65.1 & 79.8 & 69.6 & 78.4 & 85.5\\
 & 3 (QuAC) & 48.5 & 70.0 & 54.4 & 76.2 & 79.0\\
 & 4 (TrWeb) & 66.8 & 78.8 & 69.4 & 81.5 & 80.4\\
\bottomrule
\end{tabular}
}
\end{center}
\caption[caption]{\textbf{Performance of the first dataset as training progresses for text classification and question answering tasks over different dataset orders and models.}  Where applicable, we use $r_{\mathcal{M}}=100\%$ unless denoted otherwise. ``0 (Initial)" denotes model before training on any dataset.}
\label{tab:first_dataset_performance}
\end{table*}

\section{Catastrophic Forgetting}
\label{sec:first_task_progress}
To understand the severity of the catastrophic forgetting of different models, in Figure \ref{fig:catforget} and Table \ref{tab:first_dataset_performance}, we report the performance on the first dataset as training progresses. For example, we show results for AGNews as we train different models on  AGNews$\rightarrow$Yelp$\rightarrow$Amazon$\rightarrow$Yahoo$\rightarrow$DBpedia dataset order in lifelong learning setup. We also show the results prior to training on any dataset (denoted by ``Initial").

\end{document}